\documentclass[11pt,a4paper]{article}
\usepackage{times,latexsym}
\usepackage{url}
\usepackage[T1]{fontenc}
\usepackage{booktabs}
\usepackage{multirow}
\usepackage{graphicx}
\usepackage{amsmath}
\usepackage{amsfonts}
\usepackage{todonotes}

\usepackage[acceptedWithA]{tacl2021v1}

\title{OpenMSD: Towards Multilingual Scientific Documents 
Similarity Measurement}

\author{
  Yang Gao$^\diamond$, Ji Ma$^\diamond$, 
  Ivan Korotkov$^\diamond$, 
  Keith Hall$^\dagger$,
  Dana Alon$^\diamond$
  \and
  Don Metzler$^\diamond$
  \\
  \ \\
  $^\diamond$Google Research, $^\dagger$Sizzle AI
  \\
  \texttt{\{gaostayyang,maji,ivankr,danama,metzler\}@google.com}
  \\
  \texttt{khallbobo@gmail.com}
}

\date{}

\begin{document}
\maketitle
\begin{abstract}
We develop and evaluate
multilingual \emph{scientific documents similarity measurement}
models in this work. 
Such models can be used to find related works in different languages,
which can help multilingual researchers find and explore papers more efficiently. 
We propose the first multilingual scientific documents dataset,
\emph{Open-access Multilingual Scientific Documents} (OpenMSD),
which has 74M papers in 103 languages 
and 778M citation pairs. 
%
With OpenMSD, we pretrain science-specialized language models,
and explore different strategies to derive
``related'' 
paper pairs to fine-tune the models,
including using a mixture of \emph{citation}, \emph{co-citation}, and 
\emph{bibliographic-coupling} pairs. 
%
To further improve the models' performance for non-English papers,
we explore the use of \emph{generative language models} to  
enrich the non-English papers with English summaries.
This allows us to leverage the models’ English capabilities to 
create better representations for non-English papers. 
Our best model significantly outperforms strong baselines by 
7-16\% (in mean average precision).
\end{abstract}

\section{Introduction}
\label{section:introduction}
%
Although English is the predominant language
in scientific publications \cite{liu2017changing},
\emph{diversity and internationalization} in the 
scientific community has attracted more attention in recent years
\cite{uzuner2008multilingual,marquez2020science}.
Over 75\% of researchers use English as a foreign language 
\cite{unesco-science-report-2016}, 
and they often need to search related papers in both
their native languages and in English.
Think tanks and decision-making agencies 
also need to find related works
in different languages on the same topic, e.g., natural 
resource management and biodiversity studies, 
to ensure their analyses and decisions 
are unbiased and consider all affected countries
\cite{steigerwald2022overcoming}.
As the volume of non-English papers has
rapidly grown since 2000, steadily accounting for
5-10\% of all scientific publications
\cite{fortunato2018science,bornmann2021growth,nonen-paper-percent-2019},
the scientific community has an ever stronger need for 
multilingual \emph{scientific documents similarity measurement}
(SDSM) models, so as to help researchers
find, discover, and explore scientific publications 
in different languages more efficiently.
This paper focuses on the development and evaluation
of multilingual SDSM models.

The state-of-the-art SDSM models, e.g,
\cite{specter2020,scincl2022,aspire-2022},
use Transformer-based 
\cite{transformers-2017} text encoders
to create dense representations for the papers. 
Starting from a \emph{pretrained science-specialized
language model} (e.g., SciBERT \cite{scibert-2019}),
they fine-tune a \emph{dual encoder} model
\cite{gillick2018end} 
with \emph{contrastive learning objectives} 
\cite{contrastive-learning-2005,contrastive-learning-2018},
by using ``related'' and ``unrelated'' pairs of 
papers derived from citation-based heuristics or
\emph{graph embedding} algorithms \cite{deepwalk-2014,big-graph-2019}.
These models show promising performance on several SDSM tasks,
e.g., citation prediction and paper recommendation.
However, 
all of these SDSM models were trained with English data
(e.g., the S2ORC \cite{s2orc-2020} dataset) and hence
only work for English papers.
%


We identify three main challenges to develop 
multilingual SDSM models.
\textbf{(i)} 
There are no multilingual scientific documents
datasets to train and evaluate multilingual SDSM models.
\textbf{(ii)} 
There are no science-specialized multilingual language models. 
\textbf{(iii)} 
As the citation graphs' structures for
English and multilingual papers are
very different (e.g., non-English papers have much fewer
citation links than English papers, see \cite{di2017publish}
and Table \ref{tab:openmsd_key_stats}),
it is unclear whether the 
``related'' and ``unrelated'' pairs
extracted by the existing methods
are still effective for training multilingual SDSM models.

In this paper, we propose both data and novel methods
for the \emph{multilingual SDSM} problem.
For data, we build the 
\emph{Open-access Multilingual
Scientific Documents} (OpenMSD) dataset, which
has 74M papers 
and 778M citations. 
Key statistics of OpenMSD are 
presented in Table \ref{tab:openmsd_key_stats}.
Three SDSM tasks --
\emph{citation}, \emph{co-citation} \cite{cocitation-1973}, and 
\emph{bibliographic-coupling} \cite{bibcouple-1963} prediction --
are derived from OpenMSD.
%
%
To the best of our knowledge, OpenMSD is the first
multilingual scientific documents and citation relations dataset.
Scripts for reconstructing the OpenMSD dataset 
are available at 
\url{https://github.com/google-research/google-research/tree/master/OpenMSD}.\footnote{We did not directly release the dataset
due to copyright and license restrictions.}

\begin{table}[t]
    \centering
    \small
    \begin{tabular}{l l l}
    \toprule
    \multirow{8}{1.5cm}{Papers (74M)} &
    \#Papers  &  \\
    & \hspace*{10pt} $\bullet$  w/ abstracts & 74M \\
    & \hspace*{10pt} $\bullet$  w/ citations & 53M \\
    & \hspace*{10pt} $\bullet$  w/ full content & 38M \\
    & \hspace*{10pt} $\bullet$ in English & 65M \\
    & \#Abstract avg tokens & 288 \\
    & \#Content avg tokens & 5448 \\
    & \#Total tokens & 228B \\
    & \#Languages & 103 \\ 
    & \#Categories & 340 \\
    \midrule  
    \multirow{4}{1.5cm}{Citation Pairs (778M)}
    & \#En$\rightarrow$En & 759M \\ 
    & \#En$\rightarrow$nonEn & 6M \\ 
    & \#nonEn$\rightarrow$En & 11M \\
    & \#nonEn$\rightarrow$nonEn & 2.5M \\ 
    \bottomrule
    \end{tabular}
    \caption{Key statistics of the OpenMSD dataset.}
    \label{tab:openmsd_key_stats}
\end{table}

To develop multilingual SDSM models, 
we make explorations on three directions. 
\textbf{(i)} Since there are no 
science-specialized multilingual language models, we 
systematically explore different training objectives 
and data sources for developing such models,
and benchmark their performance on 
multilingual SDSM tasks.
\textbf{(ii)} We systematically investigate the effectiveness
and limitations of the latest SDSM models,
e.g., Specter \cite{specter2020} and 
SciNCL \cite{scincl2022}, in the 
multilingual setup, and propose new methods to enhance their
performance, e.g., use a mixture of different citation-based heuristics 
to create training examples.
%
\textbf{(iii)}  
To further improve the performance
for non-English papers, we propose to use
generative language models 
to create English summaries for non-English papers, and
concatenate the summaries to the
original (non-English) text,
so as to leverage the model's
English capabilities to create better representations for non-English papers.
Our best models significantly outperform strong baselines 
(SOTA SDSM models on translated text) by 7-16\% (in mean average precision).


\section{Related Works}
\label{section:related_works}

\paragraph{Scientific documents dataset.}
Several scientific documents
datasets have been compiled with open-access papers.
The \emph{arXiv Dataset} \cite{arxiv-dataset} contains the metadata
and PDFs of 1.7M papers, 
and the \emph{PMC Open Access Subset}
\cite{pmc-oa-2003}
contains the full contents of 8M papers from PubMed.
%
Papers on the \emph{ACL Anthology}\footnote{\url{https://aclanthology.org/}}
have also been used to build datasets, e.g., the 
\emph{ANN dataset} \cite{ann-2009} with 14K papers 
and 55K citations, the \emph{ACL ACR dataset} 
\cite{acl-acr-2008} with
11K papers,
and the upcoming \emph{ACL 60-60 dataset} \cite{acl-60-60},
which will 
provide machine translation of 10K paper titles and
abstracts randomly selected from the ACL Anthology 
from 2017-2021, and all the titles and abstracts 
from ACL 2022 (1.3K) into 60 languages.
The Allen AI Institute has 
published 
the \emph{S2ORC} dataset \cite{s2orc-2020} with 81M papers,
the \emph{SciDocs} dataset \cite{specter2020} with over 120K papers
and several categories of scientific tasks 
(classification, SDSM, recommendation),
and the \emph{S2AG} API
\cite{ss-api-2023}, which allows registered users to 
get access to 
the metadata 
(e.g., title, authors, abstract, but no full content)
of 206M papers and their citations (2.5B).
However, these datasets either lack citations
(the PubMed- and arXiv-based datasets), or 
only include English papers (the other 
mentioned datasets).

OpenMSD is the first dataset with both
multilingual papers 
and their citations.
Compared to \emph{S2ORC}, 
OpenMSD has a comparable number of papers (74M in OpenMSD vs 81M 
in S2ORC) but
3x more full-content papers (38M in OpenMSD vs 12M
in S2ORC) and 2x more citation pairs (759M 
in OpenMSD vs 381M in S2ORC).


\paragraph{Multilingual Language Models.}
With the huge success of Transformer-based 
\cite{transformers-2017}
language models for English tasks, 
a number of multilingual variants have also been proposed. 
They mostly follow the same recipe
(e.g., architecture, learning objectives, etc.)
as their original English versions, but are pretrained
with multilingual texts. 
Widely used models include encoder-only models
like mBERT \cite{mbert-2019}, XLM-R
\cite{xlmr-2020} and mDeBERTa \cite{debertav3-2021}, 
encoder-decoder models like mT5 \cite{mt5-2021} 
and mBART \cite{mbart-2020},
and decoder-only models like XGLM \cite{lin2021few} 
and BLOOM \cite{scao2022bloom}.
%
These models are benchmarked on multilingual datasets
like XTREME \cite{hu2020xtreme} and SuperGLUE
\cite{wang2019superglue}, which include a wide range of
tasks like named entity recognition, natural language inference, and 
question answering.
Some of them have also been fine-tuned to tackle downstream
science-related tasks, e.g., multilingual acronym extraction 
in scientific papers \cite{veyseh-etal-2022-macronym}, 
and multilingual bias evaluation in 
social science papers \cite{talat-etal-2022-reap}.
However, there are no pretrained multilingual
language models specialized for scientific documents, and there
are no datasets to benchmark their performance
on multilingual SDSM tasks.

\paragraph{SDSM models.}
A classic method to measure the similarity and relatedness
between papers is \emph{citation analysis}
\cite{citation-analysis-1971,citation-analysis-nico-2007}.
Based on the citation links between papers, heuristics
have been developed , e.g.,
\emph{co-citation} 
(two papers both cited by some common papers
\cite{cocitation-1973}), 
and \emph{bibliographic-coupling} 
(two papers both cite some
common papers \cite{bibcouple-1963})
to find related papers.
However, these methods do not work well for papers with sparse citation links,
e.g., papers that are newly published, in less-studied topics, 
or in non-English languages. 

Neural-based SDSM methods use different strategies 
to derive ``related'' and ``unrelated'' pairs
from the citation relations, 
and use them to fine-tune science-specialized
language models, e.g., SciBERT \cite{scibert-2019}.
For example, in the Specter \cite{specter2020} method,
if paper A cites paper B, B cites C but A does not cite
C, then (A, B) is used as a positive pair
while (A, C) is used as a negative pair.
\citet{scincl2022} proposed the \emph{Scientific documents
Neighborhood Contrastive Learning} (SciNCL) method,
which learns \emph{graph embeddings} \cite{big-graph-2019} 
from S2ORC's citation network;
with the learned embeddings, they can measure the
distance between papers on the citation graph,
and hence derive ``related'' and ``unrelated'' papers.
\citet{aspire-2022} proposed the \emph{Aspire} method, 
which considers the papers that
are co-cited in the same sentence as positive pairs,
because close proximity provides
a more precise indication of the relatedness of the papers.
Furthermore, as the citing sentences typically describe how
the co-cited papers are related, they use the citing sentences
as an additional signal to guide the model to learn on which aspects
the papers are related.
However, Aspire requires tools to parse the citations
in papers content, which are unavailable for 
multilingual scientific documents.
Also, all these methods are designed for English SDSM; 
it remains unclear whether they can be used to train
multilingual SDSM models. 

\section{The OpenMSD Dataset}
\label{section:openmsd_dataset}

\paragraph{Data sources.} The
scientific documents in OpenMSD
are extracted from two open-access data sources: 
the 202203 version of
\emph{Unpaywall snapshot}\footnote{\url{https://unpaywall.org/products/snapshot}}
(with 140M data entries)
and the 2022 April snapshot of the
\emph{CrossRef Metadata} \cite{crossref-2022}
(with 134M data entries).
%
%
Each data entry includes the title, 
Digital Object Identifier (DOI),
URLs and some additional meta information for a 
scientific publication. 
130 million papers occur in both data sources, by matching DOIs.
We scrape and clean the contents from the URLs, and 
remove the papers for which no text is scraped;
74M papers are retained, 
among which 38M have full content.
%
Citation relations in OpenMSD are
extracted from the 2022 October snapshot of the
\emph{OpenCitations} dataset 
\cite{10.1162/qss_a_00023}.
It has 1.4 billion unique citation pairs, each pair 
identified by the DOIs of its citing and cited paper. 
96\% of the DOIs appear in OpenCitations can be
found in Unpaywall or CrossRef.
We only keep paper pairs that have both the citing
and cited papers' abstract extracted (as papers without
abstracts cannot be used to train SDSM models; 
see \S\ref{section:specter_finetune}),
obtaining 778M citation pairs in the end.

\begin{figure}
    \centering
    \includegraphics[width=0.40\textwidth]{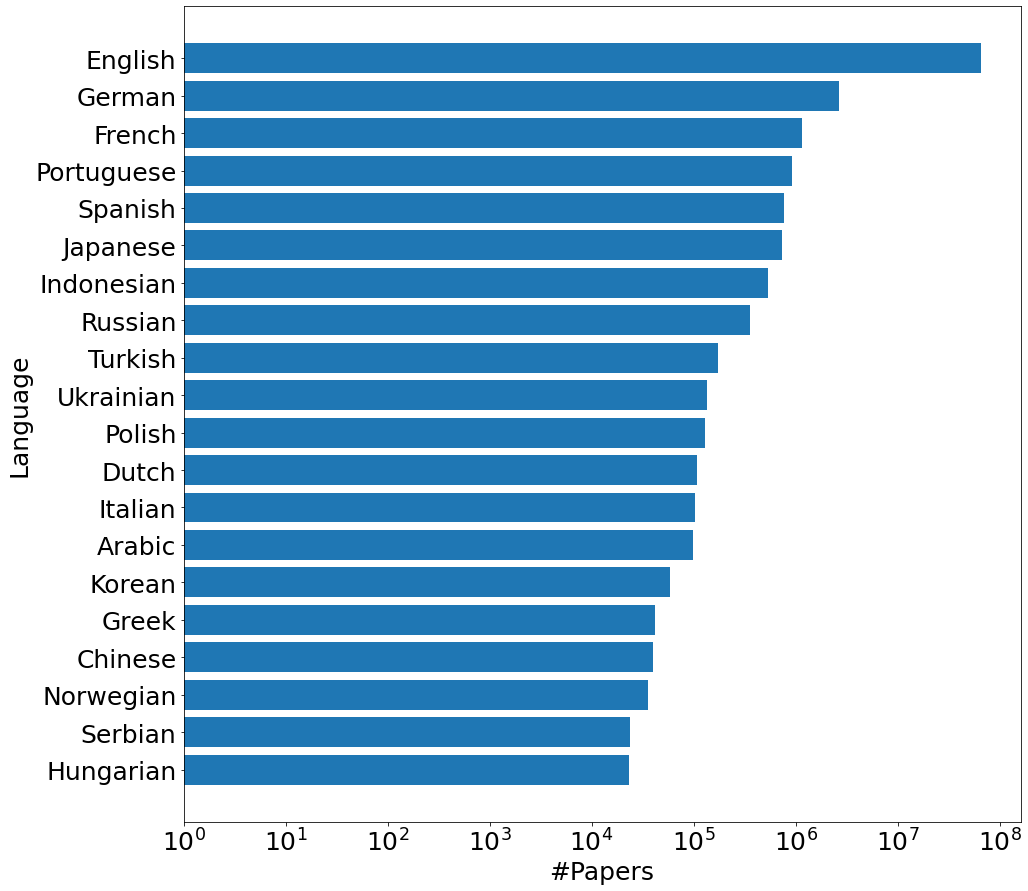}
    \caption{Top 20 languages in OpenMSD.}
    \label{fig:top_languages}
\end{figure}

\begin{figure}
    \centering
    \includegraphics[width=0.48\textwidth]{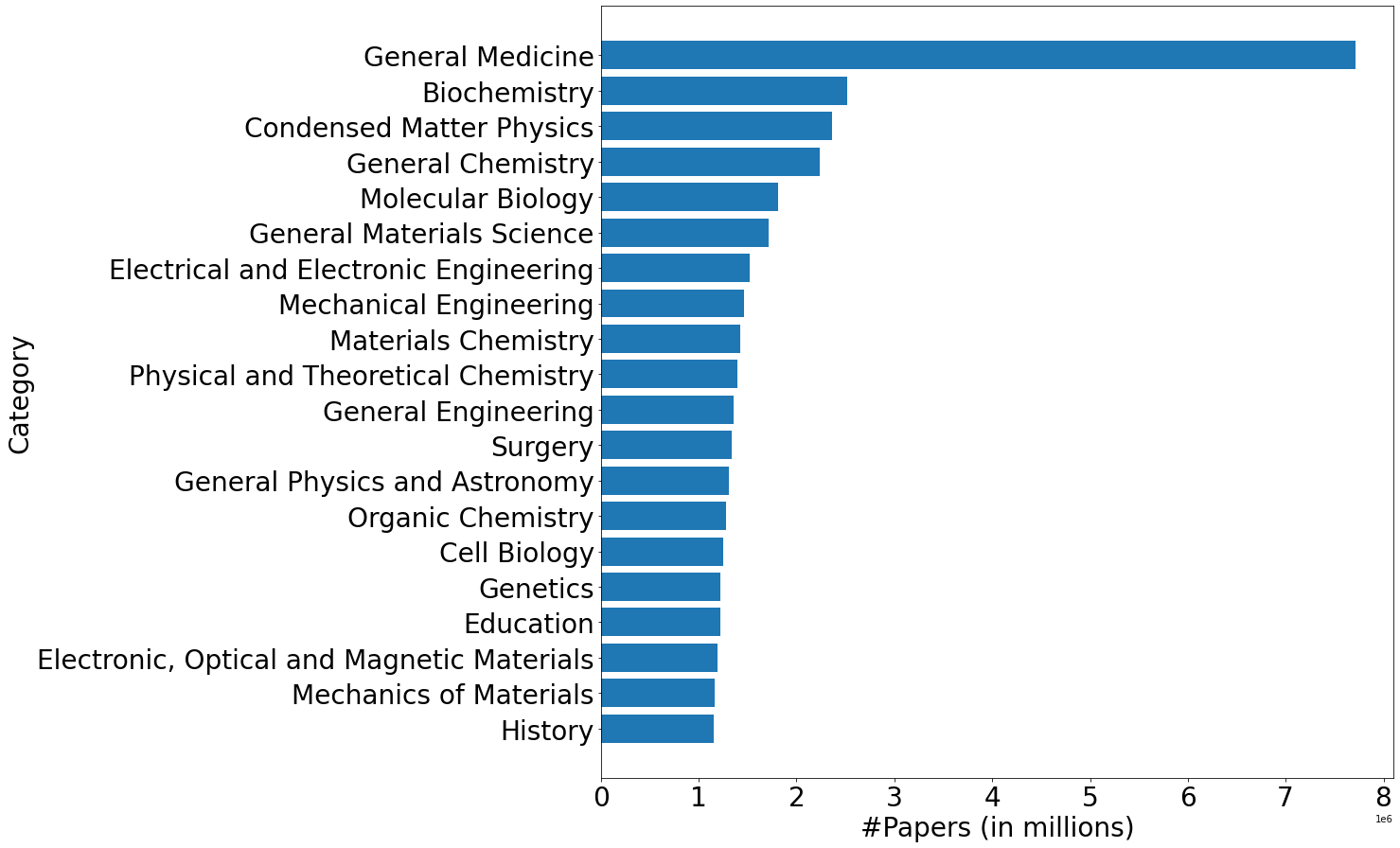}
    \caption{Top 20 categories in OpenMSD.}
    \label{fig:top20_categories}
\end{figure}

\paragraph{Languages \& Categories.}
We use \emph{cld3}\footnote{
\url{https://github.com/google/cld3}.} to detect the languages from 
papers' titles and abstracts. 
103 languages were found, with 
English (65M) being the predominant language.
%
Fig.~\ref{fig:top_languages} shows the 
the sizes of the top 20 languages.
%
Papers' category labels are extracted from CrossRef;
76\% papers have category labels, with each paper
having 1.4 category labels on average.
340 categories are found in total, and the size of the top 20 categories
are presented in Fig.~\ref{fig:top20_categories}. 
%

We note that
OpenMSD is dominated by 
English resources, which account for
88\% papers and 98\% of citation pairs 
(see Table \ref{tab:openmsd_key_stats}). 
A common strategy to
mitigate the data imbalance is to 
down-sample the English papers \cite{xlmr-2020},
but it only works well in very large datasets
like \emph{mC4} (\cite{mt5-2021},
with 6.6B pages and 6.3T tokens). 
Recent works, e.g., \cite{english-only-train-2022}, even suggest
that the English-predominance
in the training set does not necessarily hurt the 
multilingual performance, because fine-tuning multilingual models only with
English data can yield strong performance on multilingual tasks.
For these reasons, we do not perform any down-sampling 
over the English resources in OpenMSD.
Also, as scientific papers share many common 
characteristics 
regardless of their categories, we
do not manipulate the category distributions in
OpenMSD.

\paragraph{Data split.} 
To use OpenMSD to develop and evaluate 
multilingual SDSM models, 
%
we first remove all papers that do not have citation links 
with any other papers, as we cannot find their ``related'' papers;
this leaves us with 53M papers in 65 languages.
To split these papers into train and test sets,
a simple strategy is to 
randomly sample papers with a predefined ratio (e.g., 10000:1).
However, the test set built with this strategy
will be dominated by English papers and citations
(see Table \ref{tab:openmsd_key_stats})
and hence can hardly be used to evaluate 
models' performance for non-English papers. 
Also, the \emph{variance}
of the evaluation results on such test sets 
will be high, because some small languages only
have a few examples in the test set.
Furthermore, such test sets cannot be used  
to evaluate how well the multilingual models
can generalize to unseen languages, because most languages will appear 
in both train and test.

To tackle the aforementioned problems, we split the data
into train, \emph{in-distribution test} (IDT), and 
\emph{out-of-distribution test} (ODT) sets.
To create the train and IDT sets, we sample papers
in the top-30 languages 
according to their distributions in the papers pool,
and ensure that train and IDT has the same set of languages.
The remaining papers in the top-30 languages, 
together with the all the papers in the other (35) languages (around 5.5K), 
are used to build the ODT.
In addition, to avoid English-predominance in the ODT set, 
for English papers, we only keep those that are citing or cited by some
non-English papers in the ODT set.
The final train, IDT, and ODT sets have 53M, 247K and 85K papers, respectively.
The languages in each split are presented in Table \ref{tab:openmsd_splits_langs}.
With this split strategy, 
IDT can be used to evaluate the performance of
multilingual models in a more ``realistic'' setup
(as its language distribution is close to the real
language distribution of the scientific documents), while
ODT can be used to benchmark the models' performance for 
papers in non-English and unseen languages.
%
%
%
%

\begin{table}[t]
    \centering
    \small
    \begin{tabular}{p{2.3cm} p{4.5cm}}
    \toprule
    Split & Languages  \\
    \midrule
    Train (53M) \& IDT (247K) & 
    En, De, Fr, Ja, Es, Pt, Tr, Ru, 
    Id, It, Nl, Pl, Uk, Ko, Nn, Zh, 
    Cs, Hu, Lt, Da, Sv, Hr, Af, Ms, 
    Vi, Sl, Fi, Ro, Ar, Gl, 
    \\
    ODT (85K) & 
    En, Sr, De, Fr, He, Es, Pt, Ja, Fa,
    Ca, Lv, Tr, La, Sk, Su, Zh, Ru, It,
    Eu, Pl, Nl, Id, Et, Ko, Cs, Bg, 
    Hu, Sq, Is, No, Hi, Uk, Tl, Az, Af, 
    Lt, Bs, Hr, Ms, Sv, Be, Da, Eo, Mi, 
    Oc, Vi, Cy, Fi, Ia, Kk, Ku, Mk, Ro, 
    Sl, Gl, Ga, Aa, Co, Fo, Ka, El, Ky, 
    Sw, Th, Uz  
    \\
    \bottomrule
    \end{tabular}
    \caption{Languages (ISO 639-1 code) in different splits of OpenMSD,
    ordered by their sizes in each split.
    }
    \label{tab:openmsd_splits_langs}
\end{table}


With the data splits, we derive three types of
related paper pairs in each data split: 
\emph{direct citations} (DC), \emph{co-citations} (CC)
and \emph{bibliographic-coupling} (BC)
(see \S\ref{section:related_works} for definitions).
%
%
%
These relations are widely used in citation analysis
\cite{citation-analysis-nico-2007}
as indicators for related documents.
We remove pairs between papers across different splits
to avoid data leakage. Also, we remove all English to English
pairs in ODT, to make sure that ODT is focused on pairs 
involving non-English papers.
%
The numbers of mono-lingual and cross-lingual pairs of
each relation type and in each data split are
presented in Table \ref{tab:openmsd_splits_distri}.

We note that finding the related papers in IDT and ODT is
much more challenging than in existing datasets like 
SciDocs \cite{specter2020} (see \S\ref{section:related_works}).
In SciDocs' SDSM tasks (cite, co-cite, co-view and co-read), 
papers are segmented into small groups,
each with an anchor paper, five related papers to the anchor, 
and 25 randomly sampled papers; at evaluation time, models need to 
find the related papers for the anchor just from its group.
But in IDT/ODT, papers are not segmented into groups;
hence, for each paper, the models
need to find its related papers from the whole paper pool
(247K papers for IDT and 85K papers for ODT). 
We believe the setup used in IDT/ODT can better reflect
the real use cases.

\begin{table}
    \centering
    \small
    \begin{tabular}{l l  l l l }
    \toprule
    & & Train & IDT & ODT \\
    \midrule
    \multirow{4}{*}{DC} & 
    \#En$\rightarrow$En &
    759M & 229K  & 0 \\ 
    & \#En$\rightarrow$nonEn &
    6M & 2K & 3K \\
    & \#nonEn$\rightarrow$En &
    11M & 2K & 6K \\ 
    & \#nonEn$\rightarrow$nonEn &
    3M & 2K & 3K \\
    \midrule 
    \multirow{3}{*}{CC} & 
    \#En$\leftrightarrow$En & 12B & 117K & 0 \\
    & \#En$\leftrightarrow$nonEn & 208M & 2K & 1K \\
    & \#nonEn$\leftrightarrow$nonEn & 21M & 0.4K & 1K \\
    \midrule 
    \multirow{3}{*}{BC} & 
    \#En$\leftrightarrow$En & 63B & 1M & 0 \\
    & \#En$\leftrightarrow$nonEn & 1B & 11K & 7K \\
    & \#nonEn$\leftrightarrow$nonEn & 29M & 0.4K & 1K \\
    \bottomrule
    \end{tabular}
    \caption{Sizes of direct citation (DC), 
    co-citation (CC) and bibliographic-coupling (BC) pairs
    in each data split. 
    DC is a directed relation (denoted by $\rightarrow$), 
    while CC/BC are
    non-directional relations (denoted by $\leftrightarrow$).}
    \label{tab:openmsd_splits_distri}
\end{table}


\section{Pretraining Multilingual Science-Specialized Language Models}
\label{section:further_pretrain}
In this section, we develop science-specialized 
language models, which can be used as starting points
to fine-tune multilingual SDSM models.
We use \emph{mT5} 
\cite{mt5-2021} as the 
baseline and our initial checkpoint, 
as it is one of the SOTA
multilingual language models
and its encoder can be easily used in the SDSM task.
mT5 is pretrained on the \emph{mC4} dataset
with the \emph{corrupted span recovery} (CSR) 
objective.
%
%
In CSR, consecutive spans of input tokens are replaced with a
mask token and the model is trained to reconstruct
the masked-out tokens.
We use 
mT5-base because the same size of SciBERT 
is used in existing SDSM models \cite{specter2020,scincl2022}.

%

\paragraph{Further Pretraining.}
As our target task is SDSM, we aim to 
develop multilingual language models
optimized for SDSM.
Hence, besides CSR, 
we also consider the
\emph{contrastive loss} (CL) with sampled
in-batch negative \cite{sampled-in-batch-neg-2017}.
CL encourages the model to
push the positive examples closer
and the negative examples apart.
Formally, let $\{(p_i, q_i)\}_{i=1}^n$ be a training 
batch with size $n$, where $(p_i, q_i)$ is the $i$th pair
of related documents; CL is then defined as
%
\begin{align}
\label{eq:cl}
 \mathcal{L}_{CL}(\theta) = \frac{
 -\exp[ (f_\theta(p_i) \cdot f_\theta(q_i))/\tau]}{
 \sum_{j=1}^n \exp[(f_\theta(p_i) \cdot f_\theta(q_j))/\tau]},
\end{align}
\noindent where $f$ is a neural encoder parameterized by 
$\theta$,
`$\cdot$' denotes vector dot-product, and $\tau$ is the
softmax temperature.
CL has shown strong performance in both 
pretraining \cite{batch-neg-pretrain-2019} and 
fine-tuning \cite{declutr-2021,contriever-2022} 
dense representation models.
To construct the training example pairs $(p_i, q_i)$, 
we randomly extract snippets from the abstracts and 
contents of all the documents in the train set, and
the length of each snippet is between 10 and 256 mT5-sentence-piece tokens.
Snippets extracted from the same document are treated as positive example pairs.
We apply average-pooling to the output of the top transformer 
layer to get the document representation.

With the two learning objectives
(CSR and CL) and two available
datasets (mC4 and OpenMSD), we consider four
setups to further pretrain mT5, as
summarized in Table~\ref{tab:pretrain_setups}.
We use the same hyper-parameters as in mT5:
the initial learning rate is 0.001 and decayed
using the inverse square-root strategy, the 
batch size is 1K, the temperature $\tau$ is 1, 
and the models are trained with 1M steps.
0.1\% of the training data are randomly sampled and left out
as the dev set; the checkpoints with the best 
performance on the dev set are used as the final models.
All our experiments are performed on a cloud 
machine with eight TPUv3s.

\begin{table}[]
    \centering
    \small
    \begin{tabular}{p{1cm} p{0.9cm} p{0.35cm} 
    p{0.5cm} p{2.7cm}}
    \toprule 
    Name & InitCkpt & Obj. & Data & Notes \\
    \midrule 
    mT5 & Random & CSR & mC4 & Vanilla mT5 
    \\ 
    mT5CL & mT5 & CL & mC4 & 
    mT5 optimized for generic text similarity
    measurement
    \\
    mT5Sci & mT5 & CSR & OM & 
    mT5 optimized for scientific texts 
    \\
    mT5SCL & mT5 & CL & OM & 
    mT5 optimized for scientific similarity measurement 
    \\ 
    mT5CL2 & mT5CL & CL & OM & 
    Further pretrain mT5CL with 
    scientific documents \\
    \bottomrule
    \end{tabular}
    \caption{Comparing mT5 
    and its further pretrained
    models. 
    \emph{OM} stands for OpenMSD. 
    }
    \label{tab:pretrain_setups}
\end{table}

\paragraph{Results.}
We compare the mT5-based models agsint
SciBERT-base \cite{scibert-2019}.
%
%
To use SciBERT on multilingual papers, we translate
the non-English papers' titles and abstracts to
English with the Google Translate API\footnote{
\url{https://cloud.google.com/translate}.},
and use SciBERT to encode the translated text.
In line with \cite{specter2020,scincl2022},
we use the title and abstract of each document 
as input to the pretrained models, and measure the performance 
of the models by their mean average precision (MAP) and nDCG@10.

\begin{table*}[t]
    \centering
    \small
    \begin{tabular}{l  l l l l l l  l l}
    \toprule
    \multirow{2}{*}{Method} & 
    \multicolumn{2}{c}{Citation} &
    \multicolumn{2}{c}{Co-citation} &
    \multicolumn{2}{c}{Bib-couple} &
    \multicolumn{2}{c}{Average} \\
     & 
    MAP & nDCG &
    MAP & nDCG &
    MAP & nDCG &
    MAP & nDCG \\
    \midrule
    \multicolumn{9}{l}{
    \textbf{Pretrained Language Models}} \\
    SciBERT w/ translate & 0.81 & 1.13 & 0.42 & 0.90 & 0.44 & 1.38 &
    0.56 & 1.14 \\
    mT5 & 0.95 & 1.32 & 0.47 & 1.01 & 0.47 & 1.47 & 
    0.63 & 1.27 \\
    mT5Sci &  1.41 & 1.93 & 0.71 & 1.48 & 0.66 & 2.01 &
    0.93 & 1.81 \\
    mT5SCL &  1.35 & 1.86 & 0.62 & 1.33 & 0.62 & 1.92 & 
    0.86 & 1.70 \\
    mT5CL &  10.11 & 13.28 & \textbf{4.26} & \textbf{7.82} & 
    \textbf{3.55} & \textbf{8.40} & 
    \textbf{5.97} & 9.83\\
    mT5CL2 & \textbf{10.24} & \textbf{13.38} & 4.20 & 7.78 & 3.48 & 8.37 & 
    \textbf{5.97} & \textbf{9.84} \\
    \midrule
    \multicolumn{9}{l}{
    \textbf{SOTA Baselines}} \\
    \cite{specter2020} w/ translation & \textbf{17.87} & \textbf{22.74}  & \textbf{7.16} & 
    \textbf{12.23} & \textbf{6.15}  &  \textbf{12.30} &
    \textbf{10.39} & \textbf{15.76} \\
    \cite{scincl2022} w/ translation & 10.33 & 13.58 & 4.41 & 8.04 & 3.64 & 8.42 &
    6.13 & 10.01 \\
    \midrule
    \multicolumn{9}{l}{
    \textbf{Multilingual Specter (mSpt)}} \\
    mSpt$_{DC}$ & 18.52 & 23.51 & 7.38 & 12.52 & 6.23 & 12.41 &
    10.71 & 16.15 \\
    mSpt$_{CC}$ & 16.83 & 21.48 & 7.19 & 12.05 & 5.91 & 11.57 & 
    9.98 & 15.03 \\
    mSpt$_{BC}$ & 13.05 & 16.99 & 5.49 & 9.61 & 4.97 & 10.27 & 
    7.84 & 12.29 \\
    mSpt$_{DC \cup CC}$ & \textbf{19.06} & \textbf{24.15} & 
    \textbf{7.67} & \textbf{12.81} & \textbf{6.38} & \textbf{12.44} & 
    \textbf{11.04} & \textbf{16.47} \\
    mSpt$_{DC \cup BC}$ & 18.70 & 23.73 & 7.29 & 12.34 & 6.16 & 12.14 &
    10.72 & 16.07 \\
    mSpt$_{CC \cup BC}$ & 15.38 & 19.74 & 6.69 & 11.35 & 5.54 & 11.10 & 
    9.20 & 14.06 \\
    mSpt$_{DC \cup CC \cup BC}$ & 17.77 & 22.61 & 7.24 & 12.21 & 6.07 & 11.94 & 
    10.36 & 15.59 \\
    mSpt$_{DC \cap CC}$ & 18.73 & 23.77 & 7.28 & 12.41 & 6.02 & 12.31 & 
    10.68 & 16.10 \\
    mSpt$_{DC \cap BC}$ & 18.67 & 23.73 & 7.20 & 12.27 & 6.08 & 12.18 &
    10.65 & 16.06 \\
    mSpt$_{CC \cap BC}$ & 17.30 & 22.07  & 7.17 & 12.13 & 6.14 & 11.99 &
    10.20 & 15.40 \\
    mSpt$_{DC \cap CC \cap BC}$ & 18.62 & 23.66 & 7.25 & 12.39  & 6.06 & 12.15 & 
    10.64 & 16.07 \\
    \midrule 
    \multicolumn{9}{l}{
    \textbf{mSpt + Enriched Documents}} \\
    mSpt$_{DC \cup CC}$ + TopNSumm$_{64}$& 19.03 & 24.13 & 
    7.47 & 12.53 & 6.33 & 12.33 & 10.94 & 16.33 \\
    mSpt$_{DC \cup CC}$ + PaLM2Summ$_{64}$& 19.08 & 24.19 & 7.64 & 12.82 & 
    \textbf{6.39} & 12.43 & 11.04 & 16.48 \\
    mSpt$_{DC \cup CC}$ + TopNSumm$_{128}$& 19.09 & 24.21 & 7.67 & 12.88 & 
    6.38 & 12.44 & 11.05 & 16.51 \\
    mSpt$_{DC \cup CC}$ + PaLM2Summ$_{128}$& \textbf{19.22} & \textbf{24.38} & 
    \textbf{7.70} & \textbf{12.92} & 
    6.38 & \textbf{12.45} & \textbf{11.10} & \textbf{16.58} \\
    \bottomrule
    \end{tabular}
    \caption{Performance (in \%) on IDT.
    All results are averaged over 5-10 runs with different
    random seeds. }
    \label{tab:openmsd_idt_results}
\end{table*}

\begin{table*}[t]
    \centering
    \small
    \begin{tabular}{l  l l l l l l  l l}
    \toprule
    \multirow{2}{*}{Method} & 
    \multicolumn{2}{c}{Citation} &
    \multicolumn{2}{c}{Co-citation} &
    \multicolumn{2}{c}{Bib-couple} &
    \multicolumn{2}{c}{Average} \\
     & 
    MAP & nDCG &
    MAP & nDCG &
    MAP & nDCG &
    MAP & nDCG \\
    \midrule
    \multicolumn{9}{l}{
    \textbf{Pretrained Language Models}} \\
    SciBERT w/ translate & 
    1.53 & 1.95 & 0.81 & 1.41 & 
    0.40 & 0.62 & 0.91 & 1.33 \\
    mT5 & 1.62 & 2.07  & 0.99 & 1.73 & 0.46 & 0.72
    & 1.02 & 1.51 \\
    mT5Sci & 1.97 & 2.48 & 1.34 & 2.13 & 0.59 & 0.94 
    & 1.30 & 1.85 \\
    mT5SCL & 2.02 & 2.52 & 1.27 & 2.05 & 0.56 & 0.85 
    & 1.28 & 1.81 \\
    mT5CL & \textbf{7.83} & \textbf{9.51} & 
    \textbf{4.04} & \textbf{5.92} & 1.61 & 2.55 
    & 4.49 & 5.99 \\
    mT5CL2 & 7.81 & 9.45 & 4.01 & 5.81 & 
    \textbf{1.77} & \textbf{2.75}
    & \textbf{4.53} & \textbf{6.00} \\
    \midrule
    \multicolumn{9}{l}{
    \textbf{SOTA Baselines}} \\
    \cite{specter2020} w/ translation &
    \textbf{16.54} & \textbf{19.91} & \textbf{6.89} & \textbf{9.58} & 
    \textbf{3.47} & \textbf{5.02} & \textbf{8.97} & \textbf{11.50} \\
    \cite{scincl2022} w/ translation & 7.92 & 9.62 & 4.11 & 6.00 &
    1.76 & 2.71 & 4.60 & 6.11 \\
    \midrule
    \multicolumn{9}{l}{
    \textbf{Multilingual Specter (mSpt)}} \\
    mSpt$_{DC}$ & 17.64 & 21.15 & 7.15 & 9.89 & 
    3.70 & 5.33 & 9.50 & 12.12 \\
    mSpt$_{CC}$ & 15.21 & 18.39 & 6.41 & 8.92 & 
    3.22 & 4.71 & 8.28 & 10.67 \\
    mSpt$_{BC}$ & 11.72 & 14.35 & 5.08 & 7.28 & 
    3.00 & 4.33 & 6.60 & 8.65 \\
    mSpt$_{DC \cup CC}$ & \textbf{18.03} & \textbf{21.63} & 
    \textbf{7.42} & \textbf{10.11} &
    3.65 & 5.26 & \textbf{9.70} & \textbf{12.33} \\
    mSpt$_{DC \cup BC}$ & 17.77 & 21.35  & 7.09 & 9.69 &
    \textbf{3.73} & 5.28 & 9.53 & 12.11 \\
    mSpt$_{CC \cup BC}$ & 14.32 & 17.32 & 6.13 & 8.48 &
    3.22 & 4.70 & 7.89 & 10.17 \\
    mSpt$_{DC \cup CC \cup BC}$ & 
    17.03 & 20.40 & 6.75 & 9.34 & 
    3.63 & 5.20 & 9.14 & 11.65 \\
    mSpt$_{DC \cap CC}$ & 
    17.29 & 20.84 & 6.85 & 9.55 & 
    3.51 & 5.08 & 9.22 & 11.82 \\
    mSpt$_{DC \cap BC}$ & 
    17.51 & 21.00 & 6.97 & 9.69 & 
    3.74 & \textbf{5.42} & 9.41 & 12.04 \\
    mSpt$_{CC \cap BC}$ & 
    15.19 & 18.24 & 6.41 & 8.75 &
    3.28 & 4.70 & 8.29 & 10.56 \\
    mSpt$_{DC \cap CC \cap BC}$ & 16.87 & 20.16 & 6.59 & 9.37 & 
    3.53 & 5.10 & 9.00 & 11.54 \\
    \midrule 
    \multicolumn{9}{l}{
    \textbf{mSpt + Enriched Documents}} \\
    
    mSpt$_{DC \cup CC}$ + TopNSumm$_{64}$& 18.30 & 22.02 & 7.23 & 9.87 & 
    3.74 & 5.35 & 9.76 & 12.41 \\
    mSpt$_{DC \cup CC}$ + PaLM2Summ$_{64}$ & 18.85 & 22.68 & 7.29 & 9.97 & 
    3.97 & 5.68 & 10.04 & 12.78 \\
    mSpt$_{DC \cup CC}$ + TopNSumm$_{128}$& 18.55 & 22.31 & 7.23 & 9.88 &
    3.99 & 5.72 & 9.92 & 12.64 \\
    mSpt$_{DC \cup CC}$ + PaLM2Summ$_{128}$ & \textbf{19.46} & \textbf{23.40} & 
    \textbf{7.64} & \textbf{10.53} & 
    \textbf{4.05} & \textbf{5.81} & \textbf{10.38} & \textbf{13.24} \\
    \bottomrule
    \end{tabular}
    \caption{Performance (in \%) on ODT.
    All results are averaged over 5-10 runs with different
    random seeds. }
    \label{tab:openmsd_odt_results}
\end{table*}

Results on the OpenMSD's IDT and ODT sets
are presented in the top blocks in 
Table \ref{tab:openmsd_idt_results} and \ref{tab:openmsd_odt_results},
respectively.\footnote{We 
also test all models on SciDocs; the results are
presented in Table \ref{tab:scidocs_results} in Appendix 
\ref{appendix:a}.}
First, we find that vanilla mT5 outperforms SciBERT,
and we believe it is mainly because mT5 is trained with
more data: SciBERT was pretrained with 3.17B tokens,
while mT5 was pretrained with 6.3T tokens. 
Second, all further pretrained mT5-based models outperform
the vanilla mT5, suggesting that all the considered
data-objective combinations can benefit mT5's 
performance on the downstream SDSM tasks.
Third, we find that the CL objective only works well with
large data; this is reflected by the large performance
boost from mT5 to mT5CL (which uses the large mC4 data) 
and the relatively small improvement from mT5 to mT5SCL
(which uses the smaller OpenMSD data).
The improvement from mT5CL to mT5CL2 is mostly negligible,
which also suggests that the model fails to learn much 
from the second round of CL training with the (relatively small) OpenMSD data.
%
%
Given that said, mT5CL2 is still the model with the 
best average performance, and hence
we will use it as the initial checkpoint to
fine-tune our SDSM models
in the remainder of the paper,

\section{
Multilingual Specter Models}
\label{section:specter_finetune}
Specter \cite{specter2020} is the first 
Transformer-based method specialized for
English scientific NLP tasks, including SDSM. 
It uses the \emph{triplet hinge loss} to fine-tune SciBERT 
\cite{scibert-2019}. Formally,
given a triplet 
$(p_i, q_i^+, q_i^-)$, where $p_i$ is the anchor
paper, $q_i^+$ the positive example to the 
anchor, and $q_i^-$ the negative example,
the loss function is 
\begin{align}
\nonumber
\mathcal{L}_{TL}(\theta) & = 
\max\{0,   
[sim(f_\theta(p_i), f_\theta(q_i^-)) \\
& -sim(f_\theta(p_i), f_\theta(q_i^+))+m]\},
\label{eq:hinge}
\end{align}
where the hyper-parameter $m$ denotes the margin,
and the training examples are derived with 
citation-based heuristics (see \S\ref{section:related_works}).

To get multilingual SDSM models, 
we use the Specter strategy to fine-tune mT5CL2
(see \S\ref{section:further_pretrain}). 
To further improve its performance,
instead of only using direct citations (DCs) to
extract positive pairs, we explore using
co-citations (CCs) and bibliographic-couples (BCs)
in addition to DCs. 
\begin{itemize}
\item 
\textbf{Use the union of DC, CC, and BC pairs.} 
For example, we can use both DC and CC pairs as
positives, denoted as $DC \cup CC$.
Because in the train set, 
the number of CC/BC pairs is much larger
than DC (see Table \ref{tab:openmsd_splits_distri}),
we down-sample the over-represented relations 
so as to have the same number of pairs from each relation type.
\item 
\textbf{Use the intersection of DC, CC, and BC pairs}. 
Suppose a paper A cites paper B and they are both
cited by another paper C, then (A, B) is 
both a DC and CC pair.
Pairs fall into more than one relation types at the same 
time may have higher similarity level,
compared to the pairs that only fall into one type
of relation.
We consider all (four) possible intersection
combinations of the relation pairs to build positive
pairs:
$DC\cap CC$,  $DC\cap BC$,  $CC\cap BC$, and
$DC \cap CC\cap BC$. 
\end{itemize}
We use the same strategy as Specter to extract the hard negatives.
But instead of using hinge loss (Eq. \eqref{eq:hinge}),
we use the the CL objective
(Eq. \eqref{eq:cl}) during fine-tuning, because
CL can contrast each positive
pair with more negative examples (as 
all other documents in the batch are used as negatives).
Research has shown that 
replacing hinge loss with CL 
can significantly improve the performance 
\cite{declutr-2021,contriever-2022},
especially when the batch size is large 
\cite{big-batch-neg-2020}.
%
%
We denote the resulting models \emph{Multilingual Specter (mSpt)},
as they are multilingual models extending 
and generalizing Specter.

\paragraph{Baselines.}
We compare the mSpt models against 
two SOTA baselines:
Specter \cite{specter2020} 
and SciNCL \cite{scincl2022}, both applied to the translated texts.
To ensure a fair comparison, we re-implement these 
models, replace SciBERT with T5CL2 (the English-version of mT5CL2, 
based on T5 \cite{t5-2020} and  further pretrained with C4 
and English papers in OpenMSD), and fine-tune it
with OpenMSD's training data using the CL objective.
%
%
%
When implementing SciNCL, we increase the graph embedding dimension
from 768 (their original setup) to 2048, as the larger dimension size yields
better performance and 2048 is the largest dimension we manage to train 
with reasonable time and resources. 
More discussions about the SciNCL 
implementations are in \S\ref{section:distillation}.
Our preliminary experiments show that the re-implemented
versions outperform the original versions by more than 
30\% in both MAP and nDCG.
%
%
We do not re-implement Aspire \cite{aspire-2022} 
because its reported performance is close to SciNCL and
it needs to use papers cited in the same sentence
as positive pairs (see \S\ref{section:related_works});
we are not aware of tools that can reliably
parse such information from multilingual papers. 

To find the optimal hyper-parameters,
we have used batch sizes 256, 512, 1K, 2K and 4K, and
initial learning rates $10^{-n}$, where $n = 1, 2, \cdots, 7$.
The inverse square-root learning rate decay strategy is used,
with decay factor $5\times 10^{-5}$, and
the minimum learning rate is set to $10^{-8}$.
We find that batch sizes $\geq 1K$ yield similar performance,
and learning rate $10^{-2}$ yields the best performance on 
the dev set.
Each model (including mSpt and the re-implemented
SOTA baselines)
is fine-tuned for up to 100K steps, in which the 
first 1.5K steps are used for warm-up. 
Checkpoints with the best performance on the dev set
are used in the end.
%
\paragraph{Results.}
The results 
on IDT and ODT are presented in Table \ref{tab:openmsd_idt_results}
and \ref{tab:openmsd_odt_results}, respectively.
%
%
%
We find that the mSpt models' performance is significantly better\footnote{We
use double-tailed t-test $p<0.05$ as the significance test throughout this paper.} 
than all pretrained models,  
suggesting that using either DC, CC or BC to fine-tune mT5CL2
can benefit the performance.\footnote{As an
additional ablation, we also implement a \emph{random baseline},
which uses random pairs as positive examples. Its performance
is consistently 0 across all tasks and metrics.}
Compared to the SOTA baselines,
on both IDT and ODT, the best mSpt model yields significantly better performance.
In particular, on IDT, the best mSpt model is around 5\% 
better than the best baseline, while on ODT the margin
is increased to 8\%,
suggesting that mSpt performs particularly better for
non-English and unseen-language papers.
Among the mSpt models,
using both DC and CC 
as positives (i.e., $DC \cup CC$) 
yields the best performance, better than using
of any of the relation types alone or in intersections.
%
We believe this is because different citation relations
have complementary characteristics; learning from a proper mixture
of relations can help the model learn from 
each relation type, yielding more robust performance even with 
out-of-distribution data. 
%
%
This finding is significant as existing works only
use DC \cite{specter2020} or CC \cite{aspire-2022} pairs
as positive training examples.


\section{The Applicability of SciNCL on Multilingual SDSM}
\label{section:distillation}
Although SciNCL \cite{scincl2022} is reported to achieve the 
SOTA performance in the (English-only) SciDocs benchmark,
our experiments in \S\ref{section:specter_finetune}
show that it significantly underperforms 
the other fine-tuned models on multilingual SDSM.
We investigate the reason in this section.

SciNCL uses \emph{graph embedding} models to derive training pairs.
They first run 
BigGraph \cite{big-graph-2019} on the citation graph 
in S2ORC, so as to learn an embedding
for each node (i.e., paper). 
With the nodes' embeddings, they use fast nearest neighbor search algorithms
(e.g., \cite{nn-search-2020}) to find the top-$K$ neighbors for each
node, and extract positive and negative nodes therefrom:
For example, for each paper, its $i$-th closest to $(i+n)$-th closest papers 
are used as positives, while its $k$-th to $(k+n)$-th closest
papers are used as (hard) negatives, where $i, k, n \in \mathbb{N}^+$ are
hyper-parameters.
With systematic hyper-parameter search, they find that
$i=20$, $k=2000$ and $n=5$ yield the best performance.
When we re-implement SciNCL, we explore some other hyper-parameters
but find the ones used in the original work yield the
best performance.

\begin{table*}[t]
    \centering
    \small
    \begin{tabular}{l | l l l l l l | l l}
    \toprule
    \multirow{2}{*}{GraphEmbd Dim.} & 
    \multicolumn{2}{c}{Citation} &
    \multicolumn{2}{c}{Co-citation} &
    \multicolumn{2}{c}{Bib-couple} &
    \multicolumn{2}{|c}{Average} \\
    &  
    MAP & nDCG &
    MAP & nDCG &
    MAP & nDCG &
    MAP & nDCG \\
    \midrule 
    128 & 1.29 & 1.98 & 0.51 & 1.17 & 2.09 & 3.66 & 1.30 & 2.27 \\
    256 & 1.33 & 2.25 & 0.75 & 1.88 & 2.79 & 4.90 & 1.62 & 3.01 \\
    512 & 2.89 & 4.48 & 1.86 & 4.60 & 4.26 & 8.05 & 3.00 & 5.71 \\
    1024 & 4.94 & 7.16 & 3.52 & 8.32 & 5.94 & 11.94 & 4.80 & 9.14 \\
    2048 & \textbf{8.78} & \textbf{11.91} & \textbf{6.32} & \textbf{13.61} & 
    \textbf{8.05} & \textbf{17.38} & \textbf{7.72} & \textbf{14.30} \\
    \bottomrule
    \end{tabular}
    \caption{Performance (in \%) of the BigGraph \cite{big-graph-2019}
    embeddings on ODT, with different dimension sizes. 
    We have also tried DeepWalk \cite{deepwalk-2014} and InstantEmbedding
    \cite{instantembedding-2020} and they yield similar performance
    and trend. The performance in IDT is in a similar trend and hence
    omitted.}
    \label{tab:graph_embd}
\end{table*}

From the analyses above, we can view the graph embedding model
as a \emph{teacher model} and SciNCL as the 
\emph{student}.
%
Hence, to understand why the student models perform poorly,
we benchmark the teacher model by running the graph embedding 
algorithm on the train set of OpenMSD,
and evaluate its performance on ODT.
Table \ref{tab:graph_embd} 
presents the performance with different graph embedding sizes.
Comparing their performance with other systems on ODT
(see Table \ref{tab:openmsd_odt_results}),
even with dimension size 2048 (the largest dimension size
we can run in reasonable time), the graph embedding's
performance is worse than all the other fine-tuned models,
and we believe this causes the poor performance of the 
student SciNCL model.
We speculate a reason for the poor performance
of the graph embedding models is that 
the citation graph in OpenMSD is highly \emph{heterogeneous}: 
For example, the citation graph is much denser in the areas with English papers
(each English paper, on average, has 12 out-going and in-going 
citation links, respectively; see Table \ref{tab:openmsd_key_stats})
than the areas with non-English papers (each non-English paper,
on average, has only one out-going and in-going citation link
in OpenMSD).
Hence, the related/unrelated pairs derived from the graph embeddings
fail to generalize well to papers in different languages.
More rigorous investigations are required to better understand the reasons,
e.g., analyzing the topological structures of the citation graphs, and
systematically comparing different graph embedding
algorithms. It is beyond the scope of this work
and we encourage future works on it.


\section{Enrich the Non-English Documents with English Summaries}
\label{section:enrich}
Because OpenMSD is dominated by English papers and pairs 
(see \S\ref{section:openmsd_dataset}), 
models trained with OpenMSD are 
exposed more to English training examples. 
We aim to leverage the model's (relatively stronger) 
English capabilities to improve its (relatively weaker)
performance on non-English documents.
To this end, inspired by the recent works on \emph{cross-lingual summarization}
\cite{xlingual-summ-2019,xlingual-summ-survey-2022}, 
we propose to create English summaries
for the non-English papers,
and concatenate the summaries to the original (non-English) text
to create \emph{enriched documents}.
%

As there are no cross-lingual scientific documents summarization
datasets or models available, we decide to use two
\emph{zero-shot} methods to generate English summaries.
\textbf{(i)} Using the English translation of the top-N tokens 
as the summary. This is a simple yet strong baseline widely used
in summarization \cite{gao2020supert,docasref-2022}.
\textbf{(ii)} Prompting 
a large generative language model to write English summaries.
We use \emph{Flan-PaLM2} \cite{palm2} (version \emph{Otter} 
on Google Cloud API\footnote{\url{https://cloud.google.com/vertex-ai}}), 
because recent work by \citet{flan-summary-2023}
suggests that even smaller Flan-tuned language models can 
generate high-quality summaries, better than their larger but non-Flan-tuned
counterparts.
The English summary is then concatenated to the original
text in the following format: 
\emph{Title: \{title\_text\}. Abstract:  
(\{English\_summary\_text\}) \{abstract\_text\}}.
Note that English papers are not augmented with any summaries.

We consider summaries with two different lengths: 64 and 128 
tokens. To get the top-N translation summaries,
we 
simply truncate the translated abstracts to the target lengths. 
To prompt Flan-PaLM2 to generate summaries,
we experiment with a few prompts and finally use two prompts
to generate the short and long summaries, respectively:
\textbf{(i)} 
\emph{Summarize the passage below with no 
more than 30 words in English.} 
\textbf{(ii)} \emph{Extract the three most important findings 
from the passage below, and translate them to English.}
We find that the model tends to generate over-length
summaries: the average token numbers of the 
summaries generated with the two prompts above are 71 and 138, 
respectively. Over-length tokens are removed to get the
final summaries.

The enriched documents are used to train and test mSpt$_{DC \cup CC}$,
the strongest variant of mSpt.
The results of the proposed method on IDT and ODT are
presented in 
Table \ref{tab:openmsd_idt_results}
and \ref{tab:openmsd_odt_results}, respectively.
Firstly, we find that using the Flan-PaLM2-generated summaries
consistently yields better performance than the top-N translation 
summaries; we believe this is because Flan-PaLM2 considers
the whole abstract when generating the summaries, and hence
its summaries are more informative and comprehensive than 
the top-N translation summaries.
Secondly, compared to the SOTA baselines, 
using the enriched documents 
significantly boosting the MAP scores by 7\% and 16\% 
in IDT and ODT, respectively.
Compared to the vanilla mSpt$_{DC \cup CC}$,
using Flan-PaLM2-generated summaries yields marginally (not 
significantly) better performance in IDT, 
and significantly better performance in ODT.
These results suggest that enriching the 
non-English papers with high-quality English summaries can
significantly improve the multilingual models' performance for papers in
non-English and unseen languages. 

\section{Conclusion}
\label{section:conclusion}
In this work, we proposed both datasets and novel methods
for the multilingual \emph{scientific documents
similarity measurement} (SDSM) problem.
For data, we built \emph{OpenMSD}, the 
first multilingual scientific documents dataset, and
derived three SDSM tasks therefrom.
For methods, we developed science-specialized 
multilingual language models optimized for the SDSM tasks,
and fine-tuned them with related
paper pairs derived from different strategies.
Unlike existing works that use either citation
\cite{specter2020} or co-citation \cite{aspire-2022}
pairs alone, we found that using the mixture of them
yields better performance.
To further improve the model's performance for non-English documents,
we explored the use of \emph{generative language models}
to enrich the non-English papers with English summaries.
Compared to SOTA baselines, our best model improves the
performance by 7-16\% in MAP.

Our dataset and methods can be applied to other tasks beyond SDSM.
For example, OpenMSD can be used
to pretrain general-purpose large language models to 
improve their performance in reasoning and science-related tasks
\cite{taylor2022galactica,medpalm-2022}.
The technique of enriching non-English documents with
English summaries can be applied to tasks
like \emph{multilingual document clustering} \cite{multilingual-doc-cluster-2008}
and \emph{cross-lingual information retrieval} \cite{cross-lingual-ir-2015}.
More generally, 
we believe it provides a novel paradigm to leverage generative models to enhance
the text similarity measurement models; we hope this work can encourage 
more research on this direction.

\bibliography{custom}
\bibliographystyle{acl_natbib}

\newpage
\appendix 

\section{Performance on SciDcos}
\label{appendix:a}

We consider four tasks in SciDocs:
\emph{cite}, \emph{co-cite}, \emph{co-view}
and \emph{co-read}. 
Each task has a pool of 30K papers, grouped
into 1K clusters. Each cluster has one anchor paper,
five positive examples 
(e.g., in the co-view task, a positive example is a
paper that is often co-viewed with the anchor)
and 25 negative examples.
However, existing evaluations on SciDocs
\cite{specter2020,scincl2022,aspire-2022}
have shown the \emph{ceiling effect}: 
their nDCG performance
are all 90\%+ and the gaps between different methods
are rather small (less than 2 percentage points).
This is because papers are organized
into clusters at test time: given an anchor paper,
models only need to find the five positives from
the cluster (30 papers). 
%
To solve this problem, we merge the paper pools
of all four tasks and ignore the clusters at test time;
i.e., the models need to rank all 120K papers
in the merged paper pool to find the five positives.
The results are presented in 
Table \ref{tab:scidocs_results}.

\begin{table*}[t]
    \centering
    \footnotesize
    \begin{tabular}{l l l l l l l l l | l l}
    \toprule
    \multirow{2}{*}{Method} & 
    \multicolumn{2}{c}{Citation} &
    \multicolumn{2}{c}{Co-citation} &
    \multicolumn{2}{c}{Co-read} &
    \multicolumn{2}{c}{Co-view} &
    \multicolumn{2}{c}{Average} 
    \\
    & 
    MAP & nDCG &
    MAP & nDCG &
    MAP & nDCG &
    MAP & nDCG 
    & MAP & nDCG 
    \\
    \midrule
    \multicolumn{1}{l}{
    \textbf{Pretrained Language Models}} \\
    SciBERT & 
    0.52 & 2.48 & 0.51 & 2.58 &
    0.91 & 3.60 & 1.21 & 5.26 &
    0.79 & 3.48 
    \\
    mT5 & 
    0.31 & 1.41  & 0.47 & 2.31 & 
    0.49 & 1.99 & 0.92 & 3.89 &
    0.55 & 2.40
    \\
    mT5Sci & 
    0.56 & 2.70 & 0.56 & 2.83 & 
    0.98 & 3.87 & 1.30 & 5.72 &
    0.85 & 3.78
    \\
    mT5SCL & 
    0.68 & 3.31 & 0.77 & 3.85 & 
    0.93 & 3.83 & 1.36 & 5.77 &
    0.94 & 4.19
    \\
    mT5CL & 
    3.67 & 12.34 & \textbf{3.95} & 12.12 &
    4.42 & 12.63 & 5.34 & 15.13 &
    4.35 & 13.06 
    \\
    mT5CL2 & 
    \textbf{4.21} & \textbf{14.53} & 3.85 & \textbf{12.71} &
    \textbf{4.69} & \textbf{13.12} & \textbf{6.23} & \textbf{16.96} &
    \textbf{4.75} & \textbf{14.33}
    \\
    \midrule
    \multicolumn{1}{l}{
    \textbf{SOTA Baselines}} \\
    \cite{specter2020} & 
    5.51 & 15.96 & \textbf{5.76} & \textbf{16.22} & 
    6.96 & \textbf{17.32} & \textbf{9.47} & \textbf{22.43} 
    & 6.93 & \textbf{17.98}
    \\ 
    \cite{scincl2022} & 
    \textbf{5.92} & \textbf{18.04} & 5.62 & 14.37 &
    \textbf{7.56} & 16.72 & 9.24 & 21.65 & 
    \textbf{7.09} & 17.70
    \\
    \midrule
    \multicolumn{1}{l}{
    \textbf{Multilingual Specter (mSpt)}} \\
    mSpt$_{DC}$ & 
    5.60 & 16.88 & 5.61 & 14.87
    & \textbf{7.06} & 16.67 & 9.16 &  22.22
    & 6.86 & 17.66
    \\
    mSpt$_{CC}$ &
    4.75 & 14.66 & 5.43 & 14.43 & 
    6.42 & 14.94 & 8.59 & 20.12 
    & 6.30 & 16.04
    \\
    mSpt$_{BC}$ &
    4.18 & 13.48 & 4.37 & 13.17 &
    5.84 & 14.35 & 7.13 &  17.56
    & 5.38 & 14.64
    \\
    mSpt$_{DC \cup CC}$ &
    5.57 & 16.75  & \textbf{5.80} & 14.57 &
    \textbf{7.06} & 16.27 & \textbf{9.23} & 22.40 
    & \textbf{6.92} & 17.50 
    \\
    mSpt$_{DC \cup BC}$ & 
    \textbf{5.86} & 17.31 & 5.68 & \textbf{15.27} &
    6.97 & 16.26 & 9.06 &  21.81
    & 6.89 & 17.66
    \\
    mSpt$_{CC \cup BC}$ & 
    4.41 & 13.62 & 5.00 & 14.06 &
    6.31 & 14.77 & 8.33 & 19.73 
    & 6.01 & 15.55 
    \\
    mSpt$_{DC \cup CC \cup BC}$ & 
    5.56 & 17.03 & 5.29 & 14.05 & 
    6.98 & 16.37 & 8.66 & 21.38 &
    6.62 & 17.21
    \\
    mSpt$_{DC \cap CC}$ & 
    5.63 & \textbf{17.55} & 5.55 & 14.89 & 
    7.00 & \textbf{16.75} & 8.92 & \textbf{22.75} 
    & 6.78 & \textbf{17.99} 
    \\
    mSpt$_{DC \cap BC}$ & 
    5.67 & 17.23 & 5.30 & 14.82 & 
    6.28 & 15.69 & 9.19 & 22.69 
    & 6.61 & 17.61 
    \\
    mSpt$_{CC \cap BC}$ & 
    4.76 & 14.72 & 5.05 & 14.65 & 
    6.65 & 15.03 & 8.16 & 20.13
    & 6.16 & 16.13 
    \\
    mSpt$_{DC \cap CC \cap BC}$ & 
    5.52 & 17.25 & 5.34 & 14.78 & 
    6.61 & 16.04 & 8.70 & 21.57 &
    6.54 & 17.41
    \\
    \midrule 
    \multicolumn{1}{l}{
    \textbf{mSpt + Enriched Documents}} \\
    mSpt$_{DC \cup CC}$ + TopNSumm$_{64}$  & 
    5.40 & 16.36 & \textbf{5.79} & \textbf{15.84} &
    7.31 & 16.69 & 9.32 & 22.08 &
    6.96 & 17.74
    \\
    mSpt$_{DC \cup CC}$ + PaLM2Summ$_{64}$  & 
    5.41 & \textbf{16.40} & 5.75 & 15.80 & 
    7.26 & 16.74 & 9.33 & 22.07 &
    6.94 & 17.75 
    \\
    mSpt$_{DC \cup CC}$ + TopNSumm$_{128}$ &
    5.49 & 16.38 & 5.77 & 15.73 & 
    7.44 & 16.94 & \textbf{9.35} & 22.38 &
    7.01 & 17.86
    \\
    mSpt$_{DC \cup CC}$ + PaLM2Summ$_{128}$ &
    \textbf{5.58} & 16.37 & \textbf{5.79} & 15.70 & 
    \textbf{7.55} & \textbf{17.13} & 9.30 & \textbf{22.51}  
    & \textbf{7.06} & \textbf{17.93}  
    \\
    \bottomrule
    \end{tabular}
    \caption{
    Performance (in \%) on SciDocs (English-only).
    All results are averaged over 5-10 runs with different
    random seeds. 
    }
    \label{tab:scidocs_results}
\end{table*}

\end{document}